\documentclass[runningheads]{llncs}

\usepackage{eccv}

\usepackage{eccvabbrv}

\usepackage{graphicx}
\usepackage{booktabs}

\usepackage{url}
\usepackage{multirow}
\usepackage{colortbl}
\usepackage{xcolor}
\usepackage{enumitem}
\usepackage[table]{xcolor}

\usepackage[accsupp]{axessibility}  

\usepackage[pagebackref,breaklinks,colorlinks,citecolor=eccvblue]{hyperref}

\usepackage{orcidlink}

\begin{document}

\title{SPAR: Semantic-Pixel Self-Alignment and Adaptive Routing for Unified Multimodal Models} 

\titlerunning{Semantic-Pixel Self-Alignment and Adaptive Routing for Unified Models}

\author{Hongxiang Li\inst{1}\textsuperscript{*} \and
Hongxu Chen\inst{1}\textsuperscript{*} \and
Chenyang Zhu\inst{1} \and
Xiaoshuang Huang\inst{2} \and \\
Jiayin Cai\inst{2} \and
Xiaolong Jiang\inst{2} \and
Yao Hu\inst{2} \and
Long Chen\inst{1}\textsuperscript{$\dagger$}
}

\authorrunning{H.~Li et al.}

\institute{The Hong Kong University of Science and Technology \and
Xiaohongshu Inc. \\
\email{hlihg@connect.ust.hk, longchen@ust.hk} \\
Project page: \url{https://hkust-longgroup.github.io/SPAR/}
}

\maketitle

\begingroup
\renewcommand{\thefootnote}{*}
\footnotetext{Equal contribution.}
\renewcommand{\thefootnote}{$\dagger$}
\footnotetext{Corresponding author.}
\endgroup

\begin{abstract}
Multimodal Large Language Models (MLLMs) have achieved remarkable success in visual understanding but remain constrained in visual generation due to the fundamental feature discrepancy between semantic perception and pixel-level reconstruction.
Bridging this gap requires overcoming two core challenges: endowing semantic encoders with high-fidelity reconstruction capabilities, and effectively aligning generative models with semantic spaces without relying on external teachers.
To this end, we propose a novel unified multimodal framework featuring \textbf{S}emantic-\textbf{P}ixel self-alignment and \textbf{A}daptive \textbf{R}outing (\textbf{SPAR}). First, to reconcile semantic perception with pixel-level reconstruction, we introduce an asymmetric dual-stream unified tokenizer. A lightweight semantic stream anchors discriminative features, while a Transformer-augmented pixel stream recovers fine-grained visual details into a unified compact latent space. Second, to eliminate external dependencies, we propose a self-aligned generation paradigm that natively leverages this optimized tokenizer as an internal alignment teacher for the diffusion model. Furthermore, to facilitate flexible multimodal interaction within this unified space, we introduce Dynamic Token Routing, which enables each token to adaptively aggregate multi-layer MLLM features based on its distinct semantic demands. Extensive experiments demonstrate that SPAR establishes the state-of-the-art for unified architectures, achieving exceptional generation and reconstruction quality while preserving foundational visual understanding capabilities.

\end{abstract}

\section{Introduction}
Multimodal Large Language Models (MLLMs)~\cite{Qwen3-VL,wang2025internvl3_5,liu2023visual} have demonstrated exceptional performance across a wide range of perception tasks, such as image captioning, visual question answering, and multimodal reasoning, by aligning powerful visual semantic encoders with Large Language Models (LLMs)~\cite{touvron2023llama,qwen3}. 
While these models have become pervasive paradigms in the understanding domain, they have yet to dominate the field of visual generation.
Meanwhile, current state-of-the-art visual generation systems~\cite{labs2025flux1kontextflowmatching,flux-2-2025,li2025dispose,chen2026bi,wu2025qwen,wang2025target,wang2026coarse,wang2026lisa} still rely primarily on low-level, compact latent spaces constructed by Variational Auto-Encoders (VAEs)~\cite{vae}, upon which diffusion processes~\cite{esser2024scaling,rombach2022high} or autoregressive modeling~\cite{TiTok, VAR} are performed.
This leads to a discrepancy in feature patterns between perception and generation models.

\begin{figure}[t]
    \centering
    \includegraphics[width=1.\columnwidth]{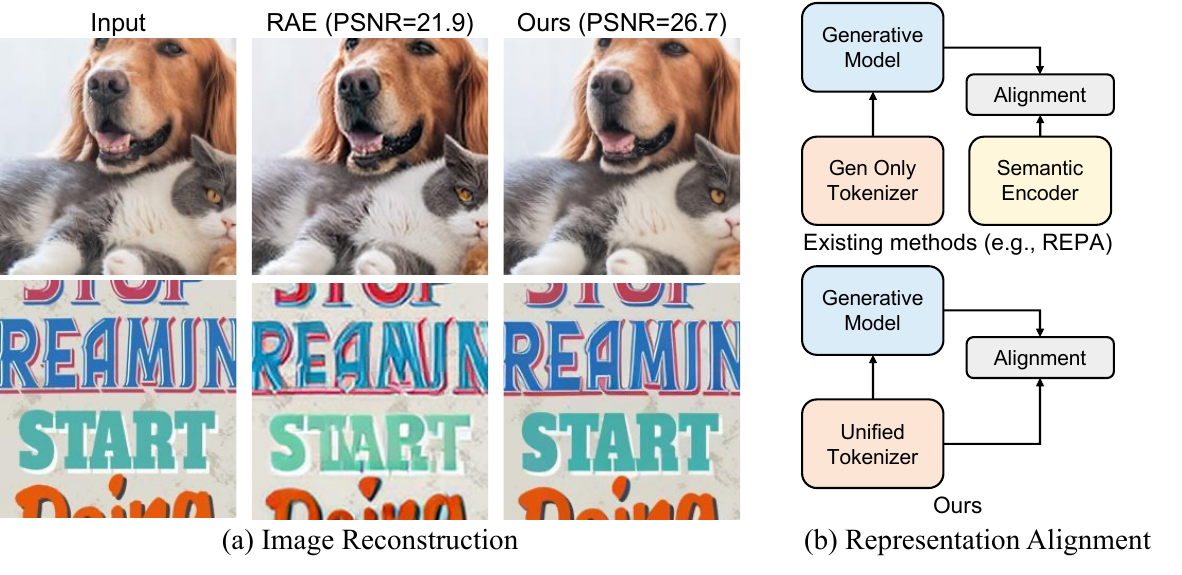}
    \vspace{-25pt}
    \caption{\textbf{(a) Image Reconstruction:} When modeling directly within the semantic representation space, existing methods suffer from lossy compression and struggle to preserve high-frequency details. In contrast, our method effectively recovers these crucial pixel-level details. \textbf{(b) Representation Alignment Paradigm:} Unlike existing approaches that rely on external {semantic encoders} to guide the generative model, our framework natively employs the unified tokenizer itself as an alignment teacher.}
    \label{fig:intro}
\end{figure}

Recent studies~\cite{yu2025repa,va-vae,rae} have explored incorporating structured semantic representations (\eg, DINOv2~\cite{oquab2023dinov2}) into generative models. By either aligning diffusion processes with these rich semantic priors or developing generative models directly upon them, these approaches have yielded significant gains in generation quality and convergence speed. 
The growing efficacy of these semantic spaces in generation reveals a promising path toward unifying visual perception and generation.
This raises \textbf{a natural and critical question}: \emph{Rather than relying on latent spaces, can we seamlessly empower powerful MLLMs with image generation capabilities by shifting the generative modeling space directly to the semantic representation space?}

However, realizing this paradigm migration is far from straightforward, as an inherent contradiction between semantic perception and pixel reconstruction needs to be overcome.
Although recent approaches~\cite{rae,va-vae} have successfully implemented diffusion models directly within the semantic representation space, they suffer from severe limitations in image reconstruction quality.
As illustrated in Figure~\ref{fig:intro}(a), RAE~\cite{rae} struggles to preserve high-frequency details, often yielding blurry artifacts and distorted structures.
This occurs because semantic encoders optimized through representation learning have feature spaces that are highly converged toward discriminative tasks. This process is essentially a lossy compression of the input image, which tends to discard texture and structural details that are irrelevant to perception. But these details may be crucial for pixel reconstruction. 
Consequently, this inherently weak reconstruction capability severely hinders the model's synthesis quality when scaled to text-to-image generation and complex instruction-based image editing~\cite{psvae}.
Furthermore, completely fine-tuning the semantic encoder to recover these lost pixel-level details would inevitably trigger catastrophic forgetting, destroying its original understanding abilities.
Therefore, {\emph{how to endow the semantic encoder with high-quality pixel reconstruction capability without compromising its original understanding ability}} remains the core bottleneck for migrating the generative modeling space to the semantic representation space.

Moreover, another critical challenge arises when effectively aligning representations to the generative model during unified model training.
As shown in Figure~\ref{fig:intro}(b), existing representation-aligned generation methods~\cite{yu2025repa,va-vae} typically bridge this gap by enforcing the diffusion model to align with external visual encoders~\cite{oquab2023dinov2} While this forced external alignment provides structured guidance, recent studies~\cite{flux2-vae} suggest that as data scale grows, relying on an isolated external teacher becomes sub-optimal. It risks manifold mismatches between the multimodal understanding and visual generation spaces. Consequently, {\emph{how to natively align the generative model directly with the unified semantic space, and eliminate the dependency on external representation learners}} is a key challenge for effectively training unified understanding and generation models.

To address the above challenges, we propose Semantic-Pixel self-alignment and Adaptive Routing (\textbf{SPAR}) for unified multimodal models. 
To overcome the inherent contradiction between semantic perception and pixel reconstruction, we introduce an asymmetric dual-stream self-alignment architecture into the unified tokenizer. 
The lightweight semantic stream acts as an anchor to preserve the encoder's original discriminative features, preventing catastrophic forgetting. Concurrently, the Transformer-augmented pixel stream is dedicated to mapping the high-dimensional semantic space into a compact latent space, effectively recovering the high-frequency spatial textures discarded by the encoder's lossy compression.
The two streams are fused in a compact latent space, explicitly decoupling semantic preservation from pixel reconstruction. 
Notably, the dual-stream-optimized tokenizer itself can serve as the representation alignment teacher for the diffusion model, seamlessly transferring the accumulated semantic and pixel knowledge to the generation stage without relying on any disjointed external feature model. 
Furthermore, we introduce Dynamic Token Routing (DTR) to facilitate flexible multimodal interaction within this unified space. It empowers each token to adaptively aggregate multi-layer MLLM features based on its distinct semantic role, dynamically satisfying the differentiated hierarchical feature demands of varying tokens.

In summary, our contributions are threefold:
\vspace{-5pt}
\begin{itemize}
    \item We propose a semantic-pixel dual-stream self-alignment architecture that explicitly decouples semantic preservation and pixel reconstruction through an asymmetric design, achieving high-fidelity reconstruction without degrading the encoder's understanding capability.
    \item We introduce dynamic token routing, which enables each token to adaptively determine its own multi-layer fusion weights from the MLLM to optimally harness multimodal representations for generative guidance.
    \item We present a self-alignment generation paradigm that leverages the unified tokenizer as an alignment teacher, removing the dependency on external feature models while achieving competitive results across multiple benchmarks.
\end{itemize}

\section{Related Work}
\noindent\textbf{Unified Visual Tokenizer.}
Early approaches~\cite{van2017neural,zhou2024transfusion} relied on reconstruction-focused generative models, which often lack the high-level semantics required for understanding tasks. To address this, recent works have explored unified tokenizers from various perspectives. Some attempt to discretize semantic features directly~\cite{wu2024vila,qu2025tokenflow,ma2025unitok}, but suffer from information loss inherent in quantization. Another recent methods~\cite{tang2025unilip,yue2025uniflow,sun2024generative,svg,rae} attempt to endow pre-trained semantic encoders with pixel reconstruction capabilities via self-distillation and continuous feature alignment. However, forcing a single representation stream to simultaneously accommodate abstract semantics and fine-grained spatial details inevitably leads to capacity bottlenecks and implicit performance trade-offs. To resolve this tension, we explicitly decouple these competing objectives through an unified tokenizer, coordinating visual understanding and generation.

\noindent\textbf{Representation-Guided Generation.}
Pre-trained visual representations are increasingly integrated into diffusion models to enhance semantic and structural fidelity. Recent methods typically achieve this by either explicitly aligning the generative latent space with external discriminative features~\cite{oquab2023dinov2,yu2025repa}, or directly performing diffusion modeling within the semantic space~\cite{rae}. However, these approaches face inherent limitations. As highly abstract semantic spaces naturally discard pixel details, such models often suffer from degraded reconstruction quality and off-manifold artifacts~\cite{psvae}.
Furthermore, extracting guidance features from external networks forces the generative process to rely on a representation space that is entirely isolated from the model's native visual encoder.
To address this, we introduce a self-aligned generation paradigm. By natively utilizing our explicitly decoupled tokenizer as an internal alignment objective.

\noindent\textbf{Unified Multimodal Understanding and Generation.}
The integration of visual understanding and generation within a single LLM has garnered significant attention~\cite{deng2025bagel,li2026girbench,zhu2026mokus,chen2025sharegpt4oimg,zhao2025unified,}. Early unified models~\cite{team2024chameleon, yu2024cm3leon} typically formulated both tasks as next-token prediction within a purely autoregressive framework. To leverage the superior synthesis quality of diffusion models, recent approaches propose attaching diffusion modules directly to the LLM backbone~\cite{sun2023emu, xie2024showo}. To bridge the modality gap, these architectures employ various connection mechanisms. For instance, some methods utilize learnable Q-Formers or linear projectors to align representations~\cite{chen2025blip3, wu2025openuni}, while others adopt Mixture-of-Experts (MoE) or specialized visual queries to manage task routing~\cite{deng2025bagel, tang2025unilip, lin2025uniworld}. 
Despite diverse connector designs, a commonality among these models is their reliance on static, token-agnostic feature extraction (\eg, using only the final layer or fixed fusion weights). This rigid paradigm fails to fully exploit the rich hierarchical representations within MLLMs. To address this, SPAR introduces dynamic token routing to adaptively aggregate multi-layer features for optimal generative guidance.

\section{Method}
\subsection{Semantic-Pixel Self-Aligned Unified Tokenizer}
The unified tokenizer aims to construct a visual encoding space that preserves both discriminative semantic information and pixel-level reconstruction capability. Consequently, it can serve as the visual encoder of the MLLM for understanding tasks while simultaneously providing a compact latent space for diffusion modeling.
Training pixel decoders directly on the frozen semantic encoder only yields blurry reconstruction results, while unlocking the encoder for reconstruction training triggers catastrophic forgetting.

\begin{figure}[t]
    \centering
    \includegraphics[width=1.\columnwidth]{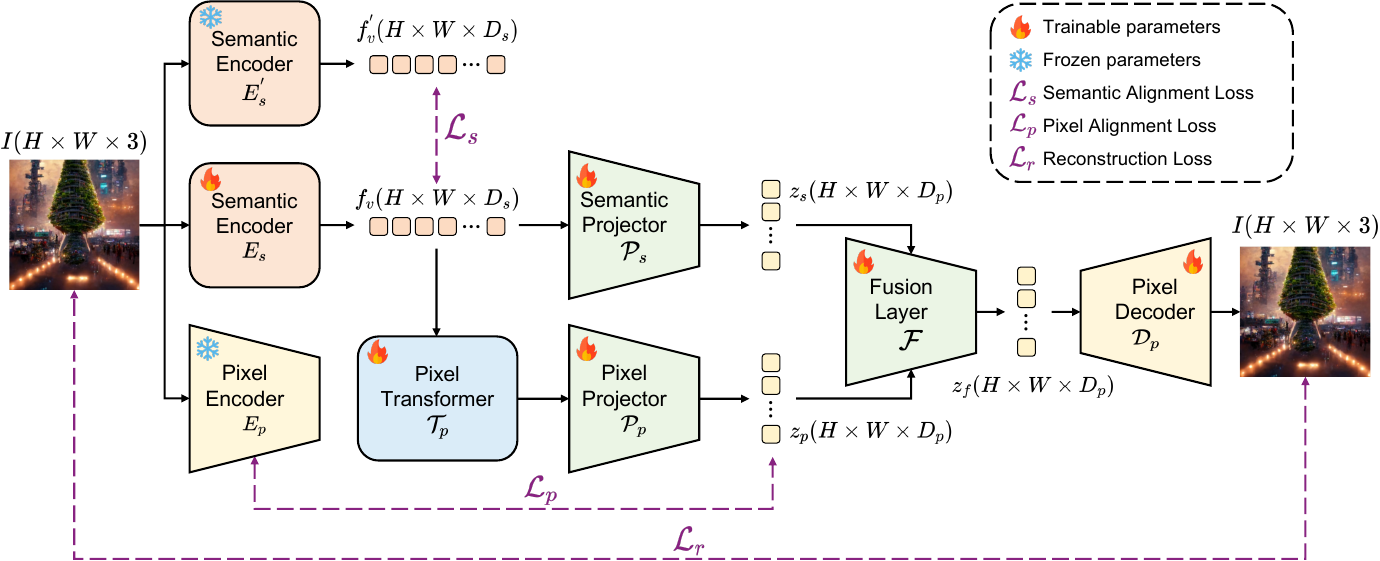}
    \vspace{-15pt}
    \caption{\textbf{Architecture of the semantic-pixel self-aligned unified tokenizer.}
    Our tokenizer design explicitly decouples semantic preservation from pixel reconstruction.
    A lightweight semantic stream maps features into a compact latent space, strictly anchored by the frozen encoder to prevent catastrophic forgetting ($\mathcal{L}_{s}$). Concurrently, a Transformer-augmented pixel stream bridges the dimensional gap by aligning with the native pixel latent space ($\mathcal{L}_{p}$) to recover high-frequency spatial details. Both streams are fused and decoded to reconstruct the image ($\mathcal{L}_{r}$), endowing the model with high-fidelity generation capabilities without compromising its discriminative understanding.
    }
    \label{fig:tokenizer}
\end{figure}

Semantic information has already been sufficiently encoded in the output features of the pre-trained encoder, and applying excessive transformations would instead perturb the original feature distribution. In contrast, the pixel-level details discarded by the encoder during representation learning require additional computational capacity for recovery. 
Based on this asymmetry, we explicitly decouple the two competing objectives of semantic preservation and pixel reconstruction as shown in~\cref{fig:tokenizer}. 
For the input image $I \in \mathbb{R} ^{H \times W \times 3}$, we construct two different streams on top of the encoder output feature $f_v =E_s(I) \in \mathbb{R}^{H^{'} \times W^{'} \times D_s}$, where $E_s$ is semantic encoder and $D_s$ is the  feature dimension.

\noindent\textbf{Semantic Stream.}
The semantic stream employs a lightweight projector $\mathcal{P}_s$ consisting of multiple residual blocks and an MLP projection layer, which maps the encoder features to the compact latent space with minimal transformation:
$\mathbf{z}_{\text{s}} = \mathcal{P}_s(f_v) \in \mathbb{R}^{H^{'} \times W^{'} \times D_p},$
where $D_p \ll D_s$ is the latent space dimension.
The residual connections ensures the output of the semantic stream to preserve the original discriminative structure of $f_v$ as much as possible, serving as an anchor for the encoder's semantic features in the latent space.

\noindent\textbf{Pixel Stream.}
A fundamental gap exists between the feature space of the semantic encoder and the compact latent space operated on by the pixel decoder. The semantic space is high-dimensional ($D_s$) and optimized for discriminative objectives, whereas the pixel latent space is low-dimensional ($D_p \ll D_s$) and tailored for reconstruction.
For the semantic encoder's output to be correctly decoded by the pixel decoder, an effective mapping from the semantic feature space to the pixel latent space must first be established. The pixel stream is designed as a learnable bridge for this purpose.
It transfers high-dimensional semantic features into the compact pixel latent space, thus allowing the fine-grained details discarded by the encoder during representation learning to be recovered within the pixel decoder's native space.
Specifically, the pixel stream is equipped with an independent Transformer encoder $\mathcal{T}_p$, which models long-range dependencies along the spatial dimension through global self-attention and transforms the semantic features into intermediate representations suitable for pixel reconstruction:
$$f_p = \mathcal{T}_p(f_v) \in \mathbb{R}^{H^{'} \times W^{'} \times D_s}.$$
The Transformer output $f_p$ is then mapped to a compact latent space compatible with the pixel decoder through residual blocks and an MLP projection layer:
$$\mathbf{z}_{\text{p}} = \mathcal{P}_p(f_p) \in \mathbb{R}^{H^{'} \times W^{'} \times D_p}.$$

\noindent\textbf{Fusion \& Decoding}.
The compact latent variables produced by the two streams are merged into a unified representation in the fusion layer $\mathcal{F}$. The semantic latent $\mathbf{z}_{\text{s}}$ and the pixel latent $\mathbf{z}_{\text{p}}$ are concatenated along the channel dimension and then mapped back to the latent space dimension by the fusion layer:
$$\mathbf{z}_f = \mathcal{F}\big([\mathbf{z}_{\text{s}};\, \mathbf{z}_{\text{p}}]\big) \in \mathbb{R}^{H^{'} \times W^{'} \times D_p},$$
where $[\cdot;\cdot]$ denotes channel concatenation. The fused $\mathbf{z}_f$ is fed into the pre-trained pixel decoder $\mathcal{D}_p$ for image reconstruction. This design allows semantic and pixel information to evolve along their respective independent optimization paths before converging in the compact latent space.

\subsection{Progressive Three-Stage Training for Unified Tokenizer}

The training of the unified tokenizer follows a progressive three-stage strategy that gradually releases model capacity, with targeted alignment losses introduced at each stage to guide the optimization direction.

\noindent\textbf{Stage I: Dual-Stream Initialization.}
The vision encoder and pixel decoder are frozen, and only the dual-stream modules are trained. The core objective of this stage is to enable the pixel stream to learn the transfer mapping from the semantic feature space to the pixel latent space. However, the semantic feature space and the pixel decoder's latent space differ fundamentally in dimensionality, distribution, and optimization objectives, making it difficult for the pixel stream to discover the correct mapping direction without explicit supervision. To this end, we introduce the \textbf{pixel alignment loss} that uses the output $\mathbf{z}'_p = E^{'}_p(I)$ of the frozen pixel encoder $E_p$ as an explicit optimization anchor and aligns the pixel stream's latent variables with it:
$$\mathcal{L}_{p} = \| \mathbf{z}_{p} - \mathbf{z}_p^{'} \|_2^2.$$
This loss first guides the pixel stream to transfer the semantic feature space onto the latent manifold of the pixel encoder, aligning its output distribution with the pixel decoder's native encoding. On this basis, the pixel stream can then effectively recover the pixel-level details discarded by the semantic encoder's lossy compression within that space. The complete training loss for this stage is:
$$\mathcal{L}_{\text{I}} = \mathcal{L}_{\text{MSE}} + \mathcal{L}_{\text{LPIPS}} + \lambda_p \mathcal{L}_{p},$$
where $\mathcal{L}_{\text{MSE}}$ represents pixel-wise reconstruction loss, and $\mathcal{L}_{\text{LPIPS}}$ represents the perceptual loss
computed using the LPIPS metric.

\noindent\textbf{Stage II: Joint Decoder Training.} The vision encoder remains frozen while the pixel decoder $\mathcal{D}_p$ is unfrozen for joint training. Only pixel reconstruction losses are used in this stage:
$$\mathcal{L}_{\text{II}} = \mathcal{L}_{\text{MSE}} + \mathcal{L}_{\text{LPIPS}}.$$
The decoder adapts to the latent distribution produced by the dual-stream fusion, maximizing reconstruction quality with a fixed encoder.

\noindent\textbf{Stage III: Encoder Fine-Tuning with Self-Distillation.}
The vision encoder is unfrozen for end-to-end training to further enhance pixel representation capability. However, unfreezing the encoder risks catastrophic forgetting: the gradients from reconstruction training may corrupt the discriminative semantic features accumulated during contrastive learning. To address this, we introduce a \textbf{semantic alignment loss} that uses the features $f'_v=E^{'}_s(I)$ of a frozen encoder to constrain the fine-tuned encoder features to maintain semantic consistency:
$$\mathcal{L}_{s} = \| f_v - f'_v \|_2^2.$$
The encoder learning rate is simultaneously decayed to $0.1\times$ the global learning rate, working together with $\mathcal{L}_{s}$ to form a dual constraint that ensures the encoder does not lose its original semantic understanding performance while acquiring stronger pixel representation capability. Additionally, a GAN discriminator is activated after a certain number of training steps to enhance the perceptual quality of reconstructed images. The complete training loss for this stage is:
$$\mathcal{L}_{\text{III}} = \mathcal{L}_{\text{MSE}} + \mathcal{L}_{\text{LPIPS}} + \lambda_s \mathcal{L}_{s} + \lambda_{\text{GAN}} \mathcal{L}_{\text{GAN}.}$$

\begin{figure}[t]
    \centering
    \includegraphics[width=1.\columnwidth]{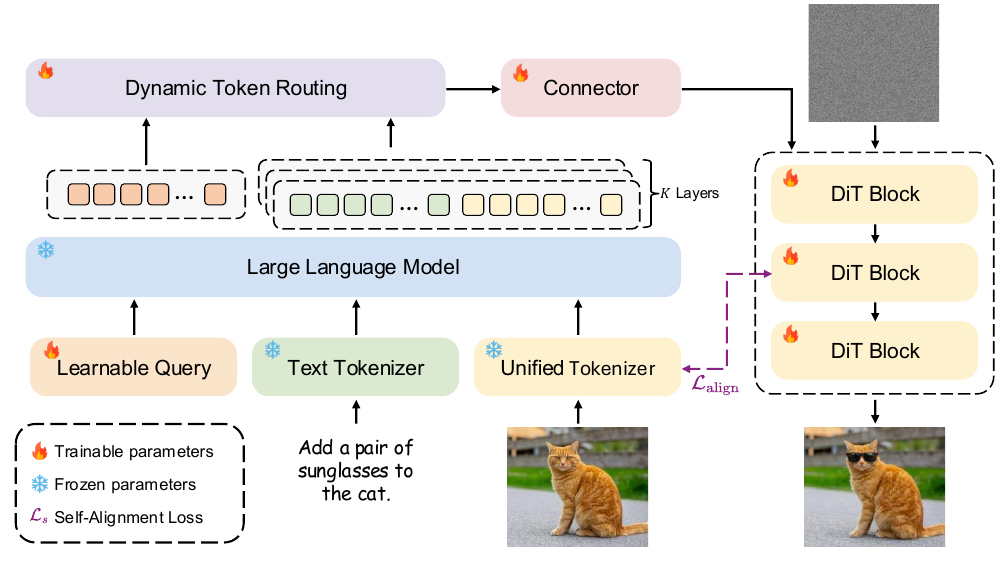}
    \vspace{-15pt}
    \caption{\textbf{Overview of the unified multimodal model.}
    The frozen MLLM processes multimodal inputs and learnable queries. The DTR adaptively aggregates multi-layer MLLM hidden states based on distinct token semantics to condition the DiT. Furthermore, the optimized tokenizer serves as an internal alignment teacher, establishing a self-alignment paradigm that eliminates reliance on external learners.
    }
    \label{fig:model}
\end{figure}

\subsection{Unified Multimodal Model}
\textbf{Conditional Signal Construction}.
To effectively bridge the multimodal reasoning capability of the MLLM with the diffusion-based generation process, we introduce a set of learnable query embeddings as implicit placeholders for the target image representation.
During training, specific image generation positions are reserved in the input token sequence, and the query embeddings are inserted at these positions by replacing the corresponding token embeddings as shown in Figure~\ref{fig:model}. The resulting composite sequence, which contains text tokens, optional reference image tokens, and query embeddings, is then processed uniformly by the MLLM's causal self-attention mechanism. The query positions located toward the end of the sequence naturally aggregate contextual information from the text instructions and visual references, thereby forming conditional representations that fuse multimodal semantics within the MLLM's hidden states.

\noindent\textbf{Dynamic Token Routing}. 
Multimodal representations within MLLMs are inherently hierarchical. Existing unified models typically extract only the last-layer hidden states or employ a token-agnostic static concatenation, completely overlooking the differentiated demands of individual tokens on hierarchical features.
To optimally harness these representations for generative guidance, we introduce Dynamic Token Routing (DTR), a mechanism that grants each token the ability to adaptively select its own multi-layer feature fusion weights according to its semantic role. Specifically, the $L$-layer Transformer of the MLLM outputs all $L+1$ hidden states (including the embedding layer output). DTR uniformly samples $K$ layers from them with sampling interval $s = \lfloor L / K \rfloor$ and collects the corresponding hidden states $\{\mathbf{H}^{(k)}\}_{k=1}^{K}$, where $\mathbf{H}^{(k)} \in \mathbb{R}^{B \times N \times D}$, stacked as $\mathbf{H} \in \mathbb{R}^{B \times N \times K \times D}$.
The routing network uses the deepest-layer features $\mathbf{H}^{(K)}$ as queries, since the top-layer features contain the most complete semantic context and are best suited for determining the role of each token. It computes the routing weights for the $i$-th token over all layers as:
$$\mathbf{w}_i = \sigma\left(\frac{g(\mathbf{H}^{(K)}_i)}{\tau}\right) \in \mathbb{R}^{K},$$
where $\sigma(\cdot)$ denotes the softmax function, $g(\cdot)$ is a lightweight routing network, and $\tau$ is a temperature coefficient. The fusion process incorporates learnable per-layer scaling parameters ${\alpha} \in \mathbb{R}^{K}$:
$$\hat{\mathbf{H}}_i = \mathbf{W}_p \left(\sum_{k=1}^{K} w_i^{(k)} \cdot \alpha^{(k)} \cdot \mathbf{H}_i^{(k)}\right),$$
where $\mathbf{W}_p$ is a linear projection.
The per-token routing weights $\mathbf{w}_i$ produced by DTR also offer a novel interpretability perspective: by visualizing the layer preference distributions of different types of tokens (image edges, texture regions, text instructions), one can intuitively reveal the model's internal decision-making mechanism during generation

\noindent\textbf{Self-Aligned Generation.}
The dual-stream-optimized unified tokenizer possesses both semantic understanding and pixel reconstruction capabilities. We leverage it directly as the representation alignment teacher for the DiT, eliminating the dependence on external pre-trained feature networks~\cite{oquab2023dinov2}. Specifically, we capture the features $f_{\text{dit}}$ at a designated intermediate layer of the DiT. These features are mapped through an alignment projection head $\phi_a$ (MLP) and aligned with the projected features $f'_v$ of the tokenizer's encoder:
$$\mathcal{L}_{\text{align}} = -\frac{1}{N}\sum_{i=1}^{N} \cos\big(\phi_a(f_{\text{dit},i}),\; f'_{v,i}\big),$$
where $\cos(\cdot, \cdot)$ denotes cosine similarity. This self-alignment paradigm seamlessly transfers the semantic and pixel knowledge accumulated during tokenizer training to the generation stage, avoiding the distribution bias that external alignment may introduce while ensuring a consistent semantic feature space shared between the understanding and generation.

\noindent\textbf{Training Strategy.} The training of the unified model has three stages:

\underline{\emph{Stage I: Connector Pre-training.}} The MLLM and DiT are frozen, and only the connector, conditional projection layer, query embeddings, and DTR routing network are trained. Generation data is used with the flow matching loss $\mathcal{L}_{\text{fm}}$ as the training objective. The goal of this stage is to enable the connector to learn the mapping from the MLLM's multimodal output to the DiT's condition space, while allowing the DTR to learn multi-layer fusion weight distribution.

\underline{\emph{Stage II: Joint Pre-training.}} The MLLM remains frozen while the connector and DiT are jointly trained. The self-alignment loss is added on top of the flow matching loss:
$\mathcal{L}_{\text{fm}} + \lambda_a \mathcal{L}_{\text{align}}.$
A mixture of generation and editing data is used for training. The DiT releases its parameters in this stage and is co-optimized with the connector, while $\mathcal{L}_{\text{align}}$ transfers the tokenizer's semantic-pixel knowledge to the DiT's intermediate representation space.

\underline{\emph{Stage III: Supervised Fine-Tuning.}} High-quality instruction tuning datasets are used to further improve generation quality and instruction-following capability. The training configuration remains the same as in Stage II.

\begin{table}[t]
\small
\centering
\caption{Comparisons of reconstruction quality on the $256 \times 256$
ImageNet 50k.}
\vspace{-10pt}
\begin{tabular}{lcccc}
\toprule
\textbf{Model} & \textbf{Ratio} & \textbf{rFID}$\downarrow$ & \textbf{PSNR}$\uparrow$ & \textbf{SSIM}$\uparrow$ \\ \midrule
\multicolumn{5}{l}{\textit{\textbf{Generative Only Tokenizer}}} \\
LlamaGen~\cite{sun2024autoregressive}      & 16 & 2.19 & 20.79 & 0.675 \\
VAR~\cite{VAR}         & 16 & 1.00 & 22.63 & 0.755 \\
Open-MAGVIT2~\cite{luo2025openmagvit2opensourceprojectdemocratizing} & 16 & 1.67 & 22.70 & 0.640 \\
RAE~\cite{rae} & 16 & 0.49 & 19.23 & 0.620 \\
SD-VAE~\cite{rombach2022high}  & 16 & 2.64 & 22.13 & 0.590 \\
DC-AE~\cite{chen2024deep} & 32 &0.69 &23.85 &0.660 \\
VA-VAE~\cite{va-vae} & 16  &0.28&\textbf{27.96} &0.790 \\
\midrule
\multicolumn{5}{l}{\textit{\textbf{Unified Tokenizer}}} \\
VILA-U~\cite{wu2024vila} & 16 & 1.80 & - & - \\
Tokenflow~\cite{qu2025tokenflow} & 16 & 1.37 & 21.41 & 0.687 \\
DualViTok~\cite{huang2025illume+} & 16 & 1.37 & 22.53 & 0.741 \\
DualToken~\cite{song2025dualtoken} &  16 &0.54 &23.56 &0.742 \\
EMU2~\cite{sun2024generative} &   14 & 3.27 & 13.49 & 0.420 \\
UniLIP~\cite{tang2025unilip} & 32 & 0.79 & 22.99 & 0.747 \\
\rowcolor[HTML]{EFEFEF}
\textbf{SPAR} & 32 & \textbf{0.27} & {26.65} & \textbf{0.856} \\
\bottomrule
\end{tabular}
\label{tab:rec}
\end{table}

\section{Experiments}

\subsection{Experimental Setups}
\noindent\textbf{Implementation Details.}
We implemented two model variants: SPAR-1B and SPAR-3B. SPAR-1B employs InternVL3-1B~\cite{zhu2025internvl3} as the MLLM backbone, which comprises InternViT as the vision encoder and Qwen2.5-0.5B~\cite{qwen2025qwen25technicalreport} as the language model, paired with SANA-0.6B~\cite{xie2024sana} as the DiT. SPAR-3B adopts InternVL3-2B as the MLLM backbone with a Qwen2.5-1.5B language model and SANA-1.6B DiT. Both variants reuse the InternViT from InternVL3 as the vision encoder and employ the decoder from DC-AE~\cite{chen2024deep} as the pixel decoder. In the asymmetric dual-stream module, the pixel stream uses a 6-layer Transformer encoder, and the semantic stream uses 3 residual blocks. DTR samples 4 layers by default with temperature coefficient $\tau=1.0$.

\noindent\textbf{Training Data.} For generation tasks, we used the combination of a 27M re-captioned data publicly released by BLIP3o~\cite{chen2025blip3}, a 5M subset of CC12M~\cite{changpinyo2021conceptual}, and 4M synthetic images from JourneyDB~\cite{sun2023journeydb}. For editing tasks, we used the GPT-Image-Edit~\cite{wang2025gpt} dataset. In the SFT stage, we employ the BLIP3o-60K and ShareGPT-4o-Image~\cite{chen2025sharegpt4oimg} high-quality datasets. Since we froze the LLM throughout training, data for understanding tasks is not required.

\begin{table}[t]
\centering
\fontsize{8.5pt}{10pt}\selectfont
\caption{Comparison with state-of-the-arts on visual understanding benchmarks.}
\vspace{-10pt}
\begin{tabular}{lccccccc}
\toprule
\textbf{Model} & \textbf{LLM Params} & \textbf{MME-P} & \textbf{MMB} & \textbf{MMMU} & \textbf{MM-Vet} & \textbf{SEED} & \textbf{MMVP} \\ \midrule
\textit{\textbf{Und. Only}} & & & & & & & \\
LLaVA-OV~\cite{li2024llava} & 1B & 1238 & 52.1 & 31.4 & 29.1 & 65.5 & - \\
InternVL3-1B~\cite{zhu2025internvl3} & 1B & 1492 & 72.6 & 43.4 & 59.5 & 71.1 & 67.3 \\
InternVL3-2B~\cite{zhu2025internvl3} & 2B & 1633 & 80.6 & 48.2 & 62.2 & 75.0 & 72.7 \\
Qwen2.5-VL-3B~\cite{Qwen2.5-VL} & 3B & - & 79.1 & 53.1 & 61.8 & - & - \\
Emu3-Chat-8B~\cite{wang2024emu3} & 8B & 1244 & 58.5 & 31.6 & 37.2 & 68.2 & 36.6 \\ \midrule
\textit{\textbf{Und. and Gen.}} & & & & & & & \\
Chameleon-7B~\cite{team2024chameleon} & 7B & - & 35.7 & 28.4 & 8.3 & - & 0.0 \\
VILA-U-7B~\cite{wu2024vila} & 7B & 1336 & 66.6 & 32.2 & 27.7 & 56.3 & 22.0 \\
MetaMorph-8B~\cite{tong2024metamorph} & 8B & - & 75.2 & 41.8 & - & - & 48.3 \\
SEED-X-13B~\cite{ge2024seed} & 13B & 1457 & 70.1 & 35.6 & 43.0 & 66.5 & - \\
Show-O-1.3B~\cite{xie2024showo} & 1.3B & 1097 & - & 26.7 & - & - & - \\
Janus-Pro-7B~\cite{chen2025janus} & 7B & 1567 & 79.2 & 41.0 & 50.0 & 72.1 & - \\
Harmon-1.5B~\cite{wu2025harmon} & 1.5B & 1155 & 65.5 & 38.9 & - & 67.1 & - \\
BAGEL-7B~\cite{deng2025bagel} & 3B & 1610 & 79.2 & 43.2 & 48.2 & - & 54.7 \\
BLIP3-o-4B~\cite{chen2025blip3} & 4B & 1528 & 78.6 & 46.6 & 60.1 & 73.8 & - \\
TokLIP-7B~\cite{lin2025toklip} & 7B & 1410 & - & 42.1 & - & 65.2 & - \\
Tar-7B~\cite{han2025vision} & 7B & 1571 & 74.4 & 39.0 & - & 73.0 & - \\
\rowcolor[HTML]{EFEFEF}
\textbf{SPAR-1B} & 1B & 1500 & 73.0 & 43.2 & 59.8 & 71.5 & 68.9\\
\rowcolor[HTML]{EFEFEF}
\textbf{SPAR-3B} & 2B & \textbf{1638} & \textbf{80.7} & \textbf{48.7} & \textbf{62.2} & \textbf{75.1} & \textbf{73.3}\\
\bottomrule
\end{tabular}
\label{tab:und}
\end{table}

\subsection{Comparisons with SOTA Methods}

\noindent\textbf{Image Reconstruction.}
Table~\ref{tab:rec} presents the comparison of image reconstruction quality on the ImageNet 50k validation set at $256 \times 256$ resolution. Compared to existing unified tokenizers, SPAR achieves state-of-the-art performance. Specifically, our model attains an rFID of 0.27, a PSNR of 26.65, and an SSIM of 0.856, significantly outperforming previous unified methods across all three metrics. Furthermore, compared with generative-only tokenizers, our model still demonstrates highly competitive reconstruction quality, outperforming strong generative baselines such as VA-VAE in SSIM ($0.856$ vs. $0.790$). 
This indicates that although our tokenizer simultaneously supports both understanding and generation within a unified framework, its reconstruction capability remains competitive with tokenizers designed exclusively for generation. Consequently, it provides a solid foundation for subsequent unified generation models built upon this representation space.

\noindent\textbf{Multimodal Understanding.}
Table~\ref{tab:und} presents the comparison of our model with recent advanced methods across various~\cite{fu2023mme,liu2024mmbench,yue2024mmmu,yu2023mm,li2024seed,tong2024eyes} visual understanding benchmarks. In our implementation, we replace the original Vision Encoder of InternVL3 with the ViT from our SPAR tokenizer. Compared to the pure understanding baseline, InternVL3, our model demonstrates superior performance across multiple benchmarks. Notably, although we unfreeze the visual encoder during training, SPAR does not suffer from any performance degradation. On the contrary, benefiting from our dual-stream self-alignment mechanism, the model achieves a noticeable performance enhancement in these understanding tasks. Furthermore, in comparison with contemporary unified models, SPAR consistently achieves superior results across all evaluated metrics.

\begin{table}[t]
\fontsize{8.5pt}{10pt}\selectfont
\centering
\caption{{Evaluation of text-to-image generation on GenEval and WISE benchmark.}}
\vspace{-10pt}
\label{tab:geneval_wise}
\begin{tabular}{lccccccc}
\toprule
\multirow{2}{*}{\textbf{Model}} & \multirow{2}{*}{\textbf{Params}} & \multicolumn{3}{c}{\textbf{GenEval}} & \multicolumn{3}{c}{\textbf{WISE}} \\
& & \textbf{Counting} & \textbf{Position} & \textbf{Overall} & \textbf{Cultural} & \textbf{Biology} & \textbf{Overall} \\
\midrule
\textit{\textbf{Gen. Only}} & & & & & & & \\
SDXL~\cite{podell2023sdxl} & 2.6B & 0.39 & 0.15 & 0.55 & 0.43 & 0.44 & 0.43 \\
FLUX.1-dev~\cite{flux2024} & 12B & 0.75 & 0.68 & 0.82 & 0.48 & 0.42 & 0.50 \\
Emu3-Gen~\cite{wang2024emu3} & 8B & 0.34 & 0.17 & 0.54 & 0.34 & 0.41 & 0.39 \\
SD3-Medium~\cite{esser2024scaling} & 2B & 0.72 & 0.33 & 0.74 & 0.42 & 0.39 & 0.42 \\
Sana-1.6B~\cite{xie2024sana} & 1.6B & 0.62 & 0.21 & 0.66 & - & - & - \\
\midrule
\textit{\textbf{Und. and Gen.}} & & & & & & & \\
VILA-U~\cite{wu2024vila} & 7B & - & - & - & 0.26 & 0.35 & 0.31 \\
TokenFlow-XL~\cite{qu2025tokenflow} & 14B & 0.41 & 0.16 & 0.55 & - & - & - \\
Janus-Pro~\cite{chen2025janus} & 7B & 0.59 & 0.79 & 0.80 & 0.30 & 0.36 & 0.35 \\
Harmon~\cite{wu2025harmon} & 3B & 0.66 & 0.74 & 0.76 & 0.38 & 0.37 & 0.41 \\
BLIP3-o-8B~\cite{chen2025blip3} & 7B & - & - & 0.84 & - & - & 0.62 \\
BAGEL~\cite{deng2025bagel} & 7B & 0.81 & 0.64 & 0.82 & 0.44 & 0.44 & 0.52 \\
OpenUni-B~\cite{wu2025openuni} & 1B & 0.74 & 0.77 & 0.84 & 0.37 & 0.39 & 0.43 \\
OpenUni-L~\cite{wu2025openuni} & 3B & 0.77 & 0.75 & 0.85 & 0.51 & 0.48 & 0.52 \\
Show-o2~\cite{xie2025show} & 7B & 0.58 & 0.52 & 0.76 & 0.33 & 0.39 & 0.39 \\
Tar~\cite{han2025vision} & 7B & 0.83 & 0.80 & 0.84 & - & - & - \\
\midrule
\rowcolor[HTML]{EFEFEF} %
\textbf{SPAR-1B} & 1B & {0.83} & 0.85 & 0.89 & 0.54 & 0.51 & 0.57 \\
\rowcolor[HTML]{EFEFEF}
\textbf{SPAR-3B} & 3B & \textbf{0.84} & \textbf{0.87} & \textbf{0.91} & \textbf{0.67} & \textbf{0.61} & \textbf{0.64} \\
\bottomrule
\end{tabular}
\end{table}

\begin{figure}[t]
    \centering
    \includegraphics[width=1.\columnwidth]{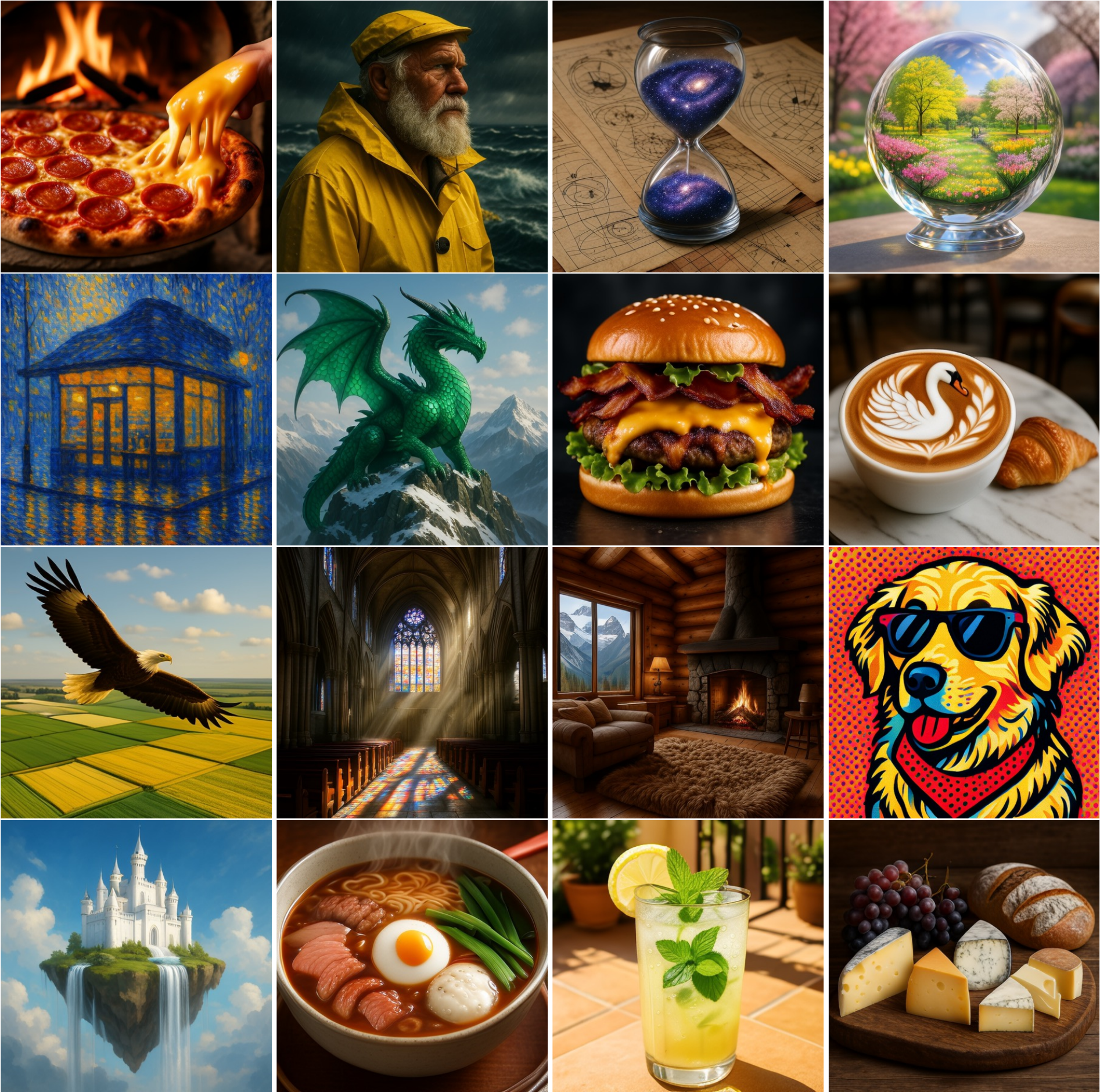}
    \vspace{-15pt}
    \caption{Qualitative results of image generation.}
    \label{fig:vis gen}
\end{figure}

\noindent\textbf{Image Generation.}
Table~\ref{tab:geneval_wise} reports the text-to-image generation results on the GenEval~\cite{ghosh2023geneval} and WISE~\cite{niu2025wise} benchmarks, and Figure~\ref{fig:vis gen} illustrates the qualitative results of our method.
We compare SPAR with two categories of methods: generation-only models and unified models that support both understanding and generation. SPAR-3B achieves an overall score of 0.91 on GenEval, ranking first among all compared methods, with the best scores of 0.84 in Counting and 0.87 in Position. Notably, as a unified model, SPAR surpasses even the much larger generation-only model, demonstrating that SPAR's representation space effectively accommodates both understanding and generation without performance degradation. On the knowledge-intensive WISE benchmark, SPAR-3B attains an overall score of 0.64, outperforming all compared methods, further validating that the rich semantic information preserved by the SPAR tokenizer benefits generation quality.

\noindent\textbf{Image Editing.}
We evaluate image editing capabilities on ImgEdit-Bench as shown in Table~\ref{tab:imgedit}. SPAR-3B achieves an overall score of 4.01, the highest among all open-source methods, closely approaching the proprietary GPT-4o (4.20). Compared with the second-best OmniGen2 (3.44), SPAR-3B yields an absolute improvement of +0.58.
Across specific editing categories, SPAR-3B achieves the best performance on various subtypes, with particularly pronounced advantages on semantically demanding tasks, indicating that the rich semantic representations within the SPAR tokenizer effectively enhance the editing model's ability to precisely comprehend and execute editing instructions.

\begin{table}[t]
\small
\centering
\caption{{Evaluation of image editing ability on ImgEdit benchmark.} }
\vspace{-10pt}
\begin{tabular}{lcccccccccc}
\toprule
\textbf{Model} & \textbf{Add} & \textbf{Adj.} & \textbf{Ext.} & \textbf{Repl.} & \textbf{Rmv.} & \textbf{Bkg.} & \textbf{Style} & \textbf{Hyb.} & \textbf{Act.} & \textbf{Overall} \\
\midrule
GPT-4o~\cite{openai2025introducing4o} & 4.61 & 4.33 & 2.9 & 4.35 & 3.66 & 4.57 & 4.93 & 3.96 & 4.89 & 4.20 \\
\midrule
MagicBrush~\cite{zhang2023magicbrush} & 2.84 & 1.58 & 1.51 & 1.97 & 1.58 & 1.75 & 2.38 & 1.62 & 1.22 & 1.90 \\
Instruct-P2P~\cite{brooks2023instructpix2pix} & 2.45 & 1.83 & 1.44 & 2.01 & 1.50 & 1.44 & 3.55 & 1.20 & 1.46 & 1.88 \\
AnyEdit~\cite{yu2025anyedit} & 3.18 & 2.95 & 1.88 & 2.47 & 2.23 & 2.24 & 2.85 & 1.56 & 2.65 & 2.45 \\
UltraEdit~\cite{zhao2024ultraedit} & 3.44 & 2.81 & 2.13 & 2.96 & 1.45 & 2.83 & 3.76 & 1.91 & 2.98 & 2.70 \\
OmniGen~\cite{xiao2025omnigen} & 3.47 & 3.04 & 1.71 & 2.94 & 2.43 & 3.21 & 4.19 & 2.24 & 3.38 & 2.96 \\
Step1X-Edit~\cite{liu2025step1x} & 3.88 & 3.14 & 1.76 & 3.40 & 2.41 & 3.16 & 4.63 & 2.64 & 2.52 & 3.06 \\
ICEdit~\cite{zhang2025enabling} & 3.58 & 3.39 & 1.73 & 3.15 & 2.93 & 3.08 & 3.84 & 2.04 & 3.68 & 3.05 \\
BAGEL~\cite{deng2025bagel} & 3.56 & 3.31 & 1.70 & 3.30 & 2.62 & 3.24 & 4.49 & 2.38 & 4.17 & 3.20 \\
UniWorld-V1~\cite{lin2025uniworld} & 3.82 & 3.64 & 2.27 & 3.47 & 3.24 & 2.99 & 4.21 & 2.96 & 2.74 & 3.26 \\
OmniGen2~\cite{wu2025omnigen2} & 3.57 & 3.06 & 1.77 & 3.74 & 3.20 & 3.57 & 4.81 & 2.52 & {4.68} & 3.44 \\
\rowcolor[HTML]{EFEFEF} \textbf{SPAR-3B} & \textbf{4.31} & \textbf{3.93} & \textbf{2.32} & \textbf{4.52} & \textbf{4.15} & \textbf{4.20} & \textbf{4.87} & \textbf{3.12} & \textbf{4.69} & \textbf{4.01} \\
\bottomrule
\end{tabular}
\label{tab:imgedit}
\end{table}

\subsection{Ablation Studies}

\noindent\textbf{Unified Tokenizer.}
Table~\ref{tab:abl_arch} investigates the contribution of each component in the unified tokenizer. Removing the pixel stream (Row b) leads to a clear reconstruction degradation (PSNR drops by 2.03, SSIM by 0.068), while the understanding metrics remain nearly unchanged, confirming that the semantic encoder alone cannot recover sufficient pixel-level details and the pixel stream is essential for bridging this gap. Removing the semantic stream (Row c) yields better reconstruction than Row~b, yet causes a catastrophic collapse in understanding (MMB drops from 73.0 to 18.4, MME-P from 1500 to 709), demonstrating that without the lightweight semantic anchor, end-to-end pixel-oriented optimization destroys the encoder's discriminative structure. The contrast between Rows~b and c validates the asymmetric design: the semantic stream preserves understanding at minimal cost, while the heavier pixel stream shoulders the reconstruction burden. Finally, freezing the encoder throughout training (Row d) preserves understanding well but results in severely degraded reconstruction (rFID 6.14, PSNR 16.26), verifying that Stage~III encoder fine-tuning with semantic self-distillation is critical for endowing the encoder with pixel representation capability.

\begin{table}[t]
\small
\centering
\vspace{-5pt}
\caption{Ablation on unified tokenizer. Evaluated on ImageNet 50k ($256 \times 256$) for reconstruction and multimodal benchmarks for understanding.}
\vspace{-10pt}
\begin{tabular}{clcccccc}
\toprule
\multirow{2}{*}{\textbf{Row}} & \multirow{2}{*}{\textbf{Setting}} & \multicolumn{3}{c}{\textbf{Reconstruction}} & \multicolumn{3}{c}{\textbf{Understanding}} \\
& & \textbf{rFID}$\downarrow$ & \textbf{PSNR}$\uparrow$ & \textbf{SSIM}$\uparrow$ & \textbf{MME-P} & \textbf{MMB} & \textbf{MMVP} \\ \midrule
(a) & Full Model & \textbf{0.27} & \textbf{26.65} & \textbf{0.856} & \textbf{1500} & \textbf{73.0} & \textbf{68.9} \\
(b) & w/o Pixel Stream & 0.31 & 24.62 & 0.788 & 1499 & 72.6 & 68.7 \\
(c) & w/o Semantic Stream & 0.29 & 25.28 & 0.804 & 709 & 18.4 & 50.0 \\
(d) & Frozen Encoder & 6.14 & 16.26 & 0.572 & 1492 & 72.6 & 67.3 \\
\bottomrule
\end{tabular}
\label{tab:abl_arch}
\end{table}

\noindent\textbf{Unified Model.}
Table~\ref{tab:abl_dtr} evaluates the two key components of the unified model. Replacing DTR with last-layer-only extraction (Row b) causes the largest overall drop, indicating that adaptively aggregating multi-layer MLLM features provides richer structural and semantic cues than relying solely on the top-layer representation. Removing the self-alignment loss $\mathcal{L}_{\text{align}}$ (Row c) also degrades all three metrics, confirming that leveraging the dual-stream tokenizer as an alignment teacher effectively transfers the accumulated semantic-pixel knowledge to the DiT and improves generation quality without any external feature network.

\begin{table}[t]
\small
\centering
\caption{Ablation on unified model training.}
\vspace{-10pt}
\begin{tabular}{clccc}
\toprule
\textbf{Row} & \textbf{Setting} & \textbf{GenEval}$\uparrow$ & \textbf{WISE}$\uparrow$ & \textbf{ImgEdit}$\uparrow$ \\ \midrule
(a) & Full Model & \textbf{0.89} & \textbf{0.57} & \textbf{3.85} \\
(b) & w/o DTR & 0.86 & 0.54 & 3.73 \\
(c) & w/o Self-Alignment $\mathcal{L}_{\text{align}}$ & 0.86 & 0.53 & 3.78 \\
\bottomrule
\end{tabular}
\label{tab:abl_dtr}
\end{table}

\section{Conclusion}
In this paper, we presented SPAR, a unified multimodal framework that addresses the fundamental feature discrepancy between semantic perception and pixel-level generation. To overcome the detail loss inherent in semantic spaces, we introduced an asymmetric dual-stream tokenizer that explicitly decouples semantic preservation from high-quality pixel reconstruction. Furthermore, we proposed dynamic token routing to adaptively harness multi-layer MLLM representations, along with a self- lignment paradigm that eliminates the reliance on external representation teachers. Extensive experiments demonstrate that SPAR achieves state-of-the-art performance across diverse benchmarks, delivering high-fidelity image generation and complex editing without degrading the pre-trained understanding capabilities.

\noindent\textbf{Acknowledgment.} This work was supported by  National Natural Science Foundation of China (NSFC) Young Scientists Fund Category B (62522216), National Natural Science Foundation of China (NSFC) Young Scientists Fund Category C (62402408), Hong Kong SAR Research Grants Council (RGC) Early Career Scheme (26208924), and Hong Kong SAR Research Grants Council (RGC) General Research Fund (16219025).

%
%
\bibliographystyle{splncs04}
\bibliography{main}
\end{document}